\pdfoutput=1

\documentclass[11pt]{article}

\usepackage[]{acl2024}
\usepackage{amssymb}
\usepackage{ragged2e}
\usepackage{times}
\usepackage{latexsym}
\usepackage{pgfplots}
\usepackage{mathrsfs}
\usepackage{array}
\newcolumntype{P}[1]{>{\centering\arraybackslash}p{#1}}
\usepackage{supertabular}
\usepackage{longtable}
\pgfplotsset{compat=1.18, width=7.7cm}
\usepackage[T1]{fontenc}
\usepackage[utf8]{inputenc}
\usepackage{subfig}

\usepackage{multirow}
\usepackage{graphicx}
\usepackage{float}
\usepackage{overpic}
\usepackage{enumitem}
\usepackage{microtype}
\usepackage{inconsolata}
\usepackage{booktabs}
\usepackage{stfloats}
\usepackage{extarrows}
\usepackage{amsmath,amssymb}   
\usepackage{proof}             
\usepackage{xcolor}
\usepackage{algorithm}
\usepackage{tikz}
\usetikzlibrary{tikzmark}
\usepackage[table, dvipsnames]{xcolor}
\usepackage{algpseudocode}
\usepackage[most]{tcolorbox}  

\colorlet{Red}{red!10!white}

\newcommand{\mytextbox}[2]{\tikzmarknode[draw=#1,thick,inner sep=2pt]{test}{\small #2}}
\definecolor{myred}{rgb}{0.7, 0.3, 0.0}
\definecolor{myblue}{HTML}{054488}
\definecolor{mygreen}{HTML}{056b34}
\newcommand{\red}[1]{\mytextbox{myred}{\textbf{\textcolor{myred}{#1}}}}

\newcommand{\green}[1]{\mytextbox{mygreen}{\textbf{\textcolor{mygreen}{#1}}}}

\newtcolorbox{mybox}{colback=white,
colframe=blue!75!black,
fonttitle=\bfseries,
title=Tool Calling Procedure,
  breakable}

\title{ToolGate: Contract-Grounded and Verified Tool Execution for LLMs}

\author{
~~Yanming Liu$^{1}$
~~Xinyue Peng$^{2}$
~~Jiannan Cao$^{3}$
~~Xinyi Wang 
~~Songhang Deng \\ \bf
~~Jintao Chen$^{1}$
~~Jianwei Yin$^{1}$
~~Xuhong Zhang$^{1}\footnotemark[1]$
 \\$^{1}$Zhejiang University $^{2}$Southeast University \\
$^{3}$Massachusetts Institute of Technology \\
   \texttt{{\{oceann24, zhangxuhong, zjuyjw, chenjintao\}@zju.edu.cn}}\\
\texttt{{xinyuepeng@seu.edu.cn, jiannan@mit.edu}}\\
}

\begin{document}

\maketitle

\renewcommand{\thefootnote}{\fnsymbol{footnote}}
\footnotetext[1]{Corresponding author.}
\renewcommand{\thefootnote}{\arabic{footnote}}

\begin{abstract}
Large Language Models (LLMs) augmented with external tools have demonstrated remarkable capabilities in complex reasoning tasks. However, existing frameworks rely heavily on natural language reasoning to determine when tools can be invoked and whether their results should be committed, lacking formal guarantees for logical safety and verifiability. We present \textbf{ToolGate}, a forward execution framework that provides logical safety guarantees and verifiable state evolution for LLM tool calling. ToolGate maintains an explicit symbolic state space as a typed key-value mapping representing trusted world information throughout the reasoning process. Each tool is formalized as a Hoare-style contract consisting of a precondition and a postcondition, where the precondition gates tool invocation by checking whether the current state satisfies the required conditions, and the postcondition determines whether the tool's result can be committed to update the state through runtime verification. Our approach guarantees that the symbolic state evolves only through verified tool executions, preventing invalid or hallucinated results from corrupting the world representation. Experimental validation demonstrates that ToolGate significantly improves the reliability and verifiability of tool-augmented LLM systems while maintaining competitive performance on complex multi-step reasoning tasks. This work establishes a foundation for building more trustworthy and debuggable AI systems that integrate language models with external tools.

\end{abstract}

\section{Introduction}
Large Language Models (LLMs) have achieved remarkable success in various reasoning tasks, particularly when augmented with external tools that enable them to interact with the real world \cite{yao2022react,gpt3,palm}. The integration of tools with LLMs has opened new possibilities for complex multi-step reasoning, where models can retrieve information, perform computations, and execute actions through API calls \cite{qin2024tool}. However, existing frameworks for LLM tool calling rely heavily on natural language reasoning to determine when tools should be invoked and whether their results should be trusted and committed to the system's understanding of the world \cite{Yang2024CanAM}. This reliance on implicit natural language reasoning creates challenges for ensuring logical safety, verifiability in tool-augmented LLM systems.

The fundamental problem lies in the lack of formal guarantees for tool invocation and result validation. Current approaches treat tool calling as a black-box process where the LLM decides based on its internal reasoning, without explicit mechanisms to verify whether the preconditions for tool invocation are satisfied or whether the tool's output meets the expected postconditions \cite{zhu2025conformity, shi2023large}. This can lead to several critical issues: tools may be called with insufficient or incorrect parameters, invalid results may be incorporated into the reasoning process, and the system's internal representation of the world state may become inconsistent or corrupted by hallucinated or erroneous tool outputs \cite{huang2025trustworthiness}. Moreover, as the number of available tools grows into the thousands, efficiently retrieving and selecting appropriate tools becomes increasingly challenging, requiring sophisticated retrieval mechanisms beyond simple keyword matching \cite{xu2024enhancing}. Recent approaches still lack a unified framework that provides formal guarantees for when tools can be safely invoked and when their results can be trusted. The absence of explicit state management and contract-based verification means that errors can propagate through the reasoning chain, making it difficult to identify and debug failures in complex multi-step tool-calling scenarios.

To address these limitations, we propose \textbf{ToolGate}, a forward execution framework that provides logical safety guarantees and verifiable state evolution for LLM tool calling. ToolGate introduces an explicit symbolic state space that maintains a typed key-value mapping representing trusted world information throughout the reasoning process. Each tool is formalized as a Hoare-style contract with a precondition that gates tool invocation and a postcondition that determines whether the tool's result can be committed to update the state. By combining Retrieval with embedding semantic search for efficient tool retrieval and hoare contract logical checks for safe tool execution, ToolGate ensures that the symbolic state evolves only through verified tool executions, preventing invalid or hallucinated results from corrupting the world representation.


\textbf{Our Contributions.} Our contributions are detailed as follows. \begin{itemize}[leftmargin=*] 
\item We present ToolGate, a novel framework that formalizes tool calling through Hoare-style contracts, providing logical safety guarantees and verifiable state evolution for LLM tool-augmented systems.
\item We introduce an explicit symbolic state space that maintains trusted world information throughout reasoning, enabling precise precondition and postcondition checking for tool invocations.
\item We demonstrate that contract-based verification significantly improves the reliability and debuggability of tool-augmented LLM systems while maintaining competitive performance on complex multi-step reasoning tasks.
\end{itemize}

\section{Related Work}
\subsection{Tool Learning of LLMs}

The integration of external tools with Large Language Models has emerged as a critical capability for extending LLM reasoning beyond text generation to real-world interactions. Early work on function calling, such as OpenAI's function calling API, enables LLMs to invoke external functions with structured parameters\cite{instructgpt}. The ReAct framework formalizes the reasoning-acting paradigm, where LLMs explicitly alternate between reasoning steps and tool invocations, demonstrating improved performance on complex multi-step reasoning tasks \cite{Yang2024CanAM}. Building on this foundation, Tool Learning has emerged as an effective paradigm for significantly expanding the capabilities of Large Language Models \cite{schick2023toolformer, qintoolllm, cikmstep}. Early research proposed that by integrating LLMs with external tools—such as program executors or search engines \cite{erdogan2024tinyagent, paranjape2023art}. To comprehensively measure performance in tool usage, researchers have introduced a series of benchmarks to systematically evaluate dimensions ranging from API selection and parameter generation quality to generalization capabilities \cite{ye2025tooleyes, patil2024gorilla, duanytool}. These techniques have been extended to multimodal tasks like GUI Agents \cite{zhang2025api, liu2025llm} and specialized domains \cite{su2025openthinkimg}. More recently, Reinforcement Learning (RL) has been incorporated into the framework to further optimize tool-learning performance \cite{qian2025toolrl, liteaching}, yielding significant results in information retrieval and dynamic reasoning. These developments demonstrate that tool-augmented LLMs are revealing vast potential for open-domain general reasoning.

\subsection{Hoare Logic and Formal Verification}

Hoare logic \cite{hoare1969axiomatic} provides a formal system for reasoning about program correctness through preconditions and postconditions. In recent years, Formal Verification and Hoare logic has been increasingly introduced into the field of deep learning to characterize and constrain the provable behaviors of neural network systems under different inputs and internal states \cite{corsi2021formal}. As deep learning models are being widely deployed in high-risk and safety-critical domains such as autonomous driving, robotics control, medical decision-making, and industrial systems, ensuring that model outputs are not only effective but also verifiable and compliant with predefined specifications has become an increasingly important problem \cite{meng2022adversarial, swaroop2024comprehensive}. In this context, the precondition–postcondition framework provided by Hoare logic is used to specify the functional, safety, or robustness properties that neural networks must satisfy under given input conditions or first-order logic \cite{yang2024harnessing, han2024folio}, and it is further combined with neural network verification and LLMs to form a unified and rigorous approach to reasoning and verification \cite{10.1145/3706598.3714113, grigorev2025verifyllm, wang2019satnet, lin2024fvel}.

\section{Methodology}
\subsection{Problem Setting and Overview}

Tool learning equips LLMs with the ability to plan, invoke, and reason over external tools. However, hallucination propagation and unreliable tool planning remain major bottlenecks, frequently leading to unstable and unreliable outcomes. To address these problems, we propose \textbf{ToolGate}, a framework integrates both probabilistic reasoning foundations and logically verifiable guarantees. It consists of a typed symbolic world state $S$ that maintains trusted information, Hoare-style logical contracts $\{P_t\}\ t\ \{Q_t\}$ for tools, and a probabilistic reasoning mechanism driven by large language models but constrained by Hoare logic.

\textbf{Problem description.} Given an input sequence $x$ and a set of available tools $T = \{t_1, t_2, \dots, t_n\}$, tool learning aims to produce an answer through:

\begin{equation}
    y = \arg\max_{y_i}  P(y_i \mid x,\; T_0 = \{t_{x_i}\})
\end{equation}

where $T_0$ represents the tools selected based on the input $x$, along with their corresponding outputs. 

\begin{figure*}[tp]
    \centering
    \includegraphics[width=1\linewidth]{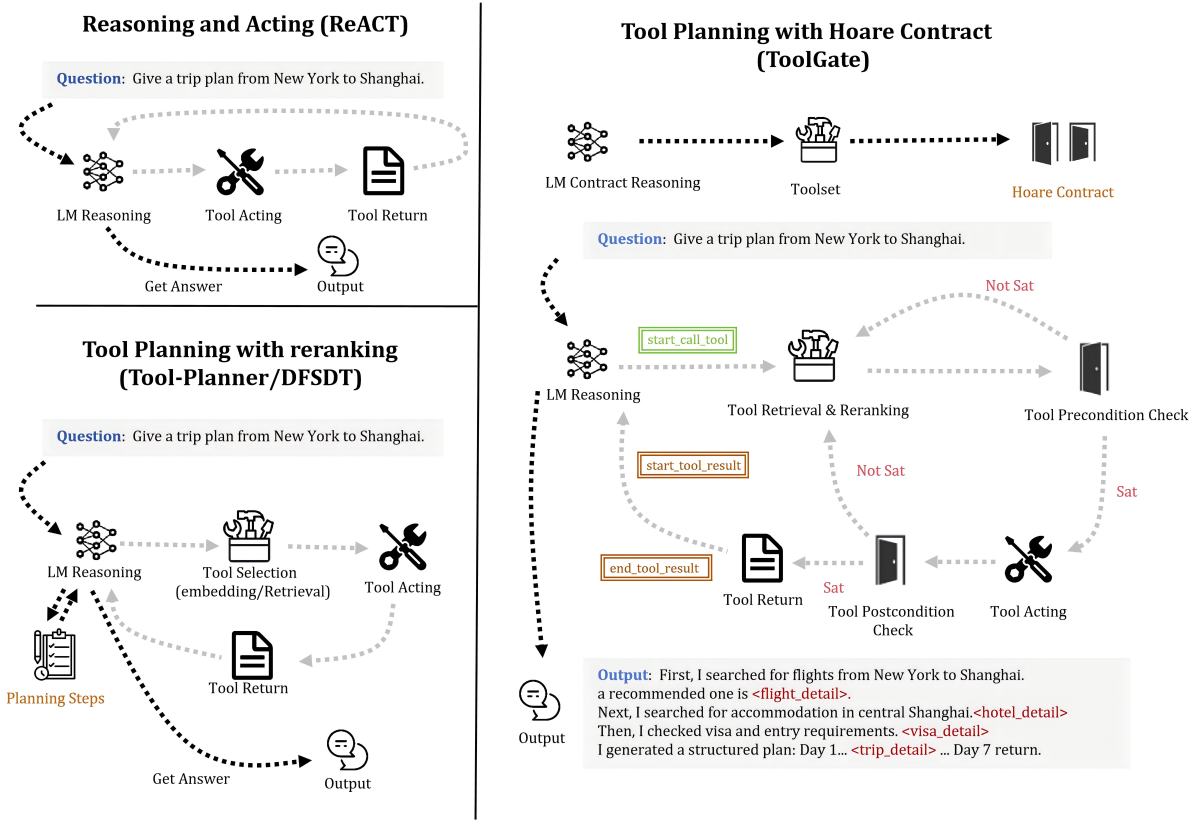}
    \caption{ToolGate framework overview. The framework is built on Hoare Logic, formalizing the tool-calling process as a sequence of constrained logical reasoning steps, and continuously maintaining a trusted state $S$ to verify the conditions for tool invocation. 
    }
    \label{fig:overview}
\end{figure*}

\subsection{Symbolic State Construction and Tool Contracts}

To ensure that tool execution is not driven merely by unstructured natural-language memory but is grounded in a verifiable and logically interpretable world model, we first construct a typed symbolic state space $\Sigma$. We maintain a trusted symbolic state $S \in \Sigma$, where each element is represented as a tuple $(k, v, \sigma)$ capturing a key, its value, and its associated type, i.e.,
\begin{equation}
\Sigma = \{(k, v, \sigma)\}
\end{equation}
This representation allows the system to explicitly encode ``what is currently known'' in a structured and inspectable manner. Verified entities, intermediate reasoning outcomes, and validated tool outputs are all written into this state space. To enforce logical consistency throughout reasoning and tool execution, we additionally define a set of logical predicates over $\Sigma$ to express existence constraints, type consistency, and semantic invariants, and we denote $S \models \varphi$ to indicate that a given symbolic state $S$ satisfies a logical condition $\varphi$. 

To prevent the model from invoking tools arbitrarily and reduce hallucinated or unconstrained execution behavior, we assign each tool $t \in \mathcal{T}$ a Hoare-style logical contract of the form
\begin{equation}
\{P_t\}\ t\ \{Q_t\}
\end{equation}
The precondition $P_t : \Sigma \rightarrow \{true,false\}$ specifies the minimal state requirements that must be satisfied for the tool to be legally callable, meaning a tool is not executable unless $S \models P_t$ holds. Meanwhile, the postcondition $Q_t : \Sigma \times R_t \rightarrow \{\text{true}, \text{false}\}.$ constrains the structural validity, typing correctness, and semantic consistency of the runtime output $r_t$, while also defining how a verified result updates the system state.

\subsection{Tool Call and Reranking}
We first treat the model’s reasoning as a process of conditional probability propagation over time. At the $k$-th step, the reasoning state is represented as
\begin{equation}
p(R_k \mid q, H, S_k)
\end{equation}
where $q$ denotes the current user query, $H$ represents the stable and externally visible dialogue history, and $S_k$ denotes the trusted symbolic state at this time. We define $R_k$ as the current reasoning trajectory, which records the intermediate reasoning content, reasoning path, and any tool results already injected before step $k$. In this formulation, $H$ tracks externally observable interaction, while $R_k$ captures the evolution of the model’s internal reasoning process, making it clear, in subsequent tool selection and state updates, which information originates from the user and which originates from internal reasoning.

Next, we turn the choice to call a tool into an endogenous stochastic decision within the reasoning process itself. Under the current information state, the model estimates:

\begin{equation}
p( \text{\green{{<start\_call\_tool>}}} \mid q, H, S_k, R_k)
\end{equation}

and this probability directly drives whether the model generates \green{{<start\_call\_tool>}}. Once this marker appears, the system enters the tool selection and execution phase; when \red{{<start\_tool\_result>}}, \red{{<end\_tool\_result>}} are later concatenated, the system exits the tool phase and returns to pure natural language answering. This design allows tool usage to be determined by the model’s uncertainty and task requirements at the moment, rather than by inflexible hand-crafted triggers, enabling smoother adaptation to scenarios where tools are sometimes necessary and sometimes unnecessary.

Based on the current query $q$, dialogue history $H$, symbolic state $S_k$, and reasoning trajectory $R_k$, we construct a tool requirement representation:

\begin{equation}
u_k = f(q, H, S_k, R_k)
\end{equation}

which provides a structured description of the present subproblem and clarifies what the system aims to achieve and what type of tool output it expects. We then treat $u_k$ as a query to retrieve from the large tool set $\mathcal{T}$, using vector embeddings jointly to extract the Top-$K$ candidate tools:
\begin{equation}
\mathcal{C}_k = \text{TopK-Retrieve}(u_k, \mathcal{T})
\end{equation}

which effectively shrinks the tool space, preserving only a small, highly relevant candidate set. 

With the candidate set $\mathcal{C}_k$, we apply a reranking model within $\mathcal{C}_k$, producing a refined ranking distribution:
\begin{equation}
p_{\text{rank}}(t \mid u_k), \quad t \in \mathcal{C}_k
\end{equation}

\subsection{Tool Contracts on Planning}

For each candidate tool $t \in \mathcal{C}_k$, we determine whether its precondition is satisfied under the current symbolic state $S_k$, using the indicator $\mathbf{1}[S_k \models P_t]$ to eliminate all tools whose prerequisites are unmet. We then renormalize the ranking distribution only over those tools whose preconditions hold, forming a logically valid execution policy:

\vspace{-0.3cm}

\begin{equation}
\begin{aligned}
& p^*(t \mid q, H, S_k, R_k) = \\
& \quad \frac{p_{\text{rank}}(t \mid u_k) \cdot \mathbf{1}[S_k \models P_t]}{\sum_{t' \in \mathcal{C}_k} p_{\text{rank}}(t' \mid u_k) \cdot \mathbf{1}[S_k \models P_{t'}]}
\end{aligned}
\end{equation}

This filtering mechanism transcends simple semantic matching by establishing formal execution admissibility; it necessitates that the current state $S_k$ satisfies the weakest precondition of the selected tool, denoted as $S_k \models wp(t, P_t)$. By embedding such deterministic constraints into the probabilistic sampling process, we ensure that the model’s trajectory remains within a logically grounded solution space rather than relying on unconstrained heuristic transitions.

We treat $p^*(t)$ as a logically constrained policy distribution and sample from it:

\begin{equation}
t^* \sim p^*(t\mid q, H, S_k, R_k)
\end{equation}

As long as a tool is both legal and meaningfully relevant, it naturally retains the chance to be explored, while its sampling probability reflects its contextual priority. Once the final tool $t^*$ is selected and invoked, it returns a result $r_t$. Before updating the system state with this output, we introduce a safety gate, a runtime contract verification process that checks whether the returned result satisfies the Hoare postcondition $Q_t$. We formalize this as a binary acceptance event $\mathcal{A}_t \in \{0,1\}$,
with conditional probability $p(\mathcal{A}_t = 1 \mid S_k, r_t, Q_t)$ and implement it as a concrete verification function:
\begin{equation}
\mathcal{A}_t =
\begin{cases}
1, & \text{if } (S_k, r_t) \models Q_t \land \text{wf}(r_t), \\
0, & \text{otherwise.}
\end{cases}
\end{equation}
Through this step, every tool output must satisfy structural validity, value range constraints, and format expectations before it can affect the global state. Only if verification passes does the symbolic state update:
\begin{equation}
S_{k+1} =
\begin{cases}
\mathsf{Update}_t(S_k,r_k), & \mathcal{A}_t = 1, \\
S_k, & \mathcal{A}_t = 0,
\end{cases}
\end{equation}
and the accepted result is injected into the subsequent reasoning trajectory $R_{k+1}$. If verification fails, the result is discarded entirely, preventing contaminated outputs from propagating and providing a clear debugging breakpoint.

Simultaneously, we inject the verified results wrapped in \red{{<start\_tool\_result>}} and \red{{<end\_tool\_result>}} tags into the subsequent reasoning trajectory $R_{k+1}$, enabling both subsequent natural language reasoning and the next round of tool selection to fully leverage this newly acquired trusted information.

Building on this foundation, we treat the entire system as a family of stochastic trajectories $\tau$, each consisting of $(S_k,R_k)$, the chosen tools $t_k$, and the acceptance events $\mathcal{A}_k$. The system performs probabilistic reasoning over all feasible execution trajectories, and the final output $y$ can be expressed via trajectory-level marginalization:

\begin{equation}
p(y \mid q,H)
=
\sum_{\tau}
p\bigl(y \mid q,H,\tau\bigr)\,
p\bigl(\tau \mid q,H\bigr)
\end{equation}

where $p(\tau \mid q,H)$ integrates all components discussed above: tool trigger probability, requirement abstraction, retrieval and ranking, contract filtering, constrained sampling, and acceptance verification.

To ensure Hoare contracts regulate not only local behavior but also the global behavior space, we impose a strict trajectory-level constraint: if any trajectory $\tau$ violates any tool precondition $P_t$ or postcondition $Q_t$ at any step, then
\begin{equation}
p(\tau \mid q,H)=0
\end{equation}

Under this formulation, reasoning and sampling proceed exclusively within a trajectory subspace that adheres to predefined contracts, thereby providing a formal logical justification for each state transition and tool execution.

\section{Experimental Setup}
\subsection{Dataset.} 
We utilize \textbf{ToolBench} \citep{qintoolllm} and \textbf{MCP-Universe} \cite{luo2025mcpuniverse} as our experimental datasets. ToolBench contains more than 16,000 APIs organized into structured tool categories, covering a wide range of functional capabilities. These settings jointly assess both local tool invocation ability and global planning robustness.

MCP-Universe reflects more realistic multi-tool environments. It aggregates diverse tools, plugins, and APIs from real-world systems covering information retrieval, automation, data processing, system operations, and task execution. We use the tools selected in ToolBench and MCP-Universe along with their official documentation, specifications, and usage descriptions to extract structured functional representations.  More dataset details are provided in Appendix~\ref{sec:appendixA}.

\begin{table*}[t]
\centering
\caption{
Main experimental results on ToolBench and MCP-Universe. 
We report Pass Rate (\%) and Win Rate (\%) for ToolBench G1, G2, and G3 tasks,
and Success Rate (\%) for three MCP-Universe subtasks.
}
\label{tab:toolgate_main_result}
\resizebox{0.98\textwidth}{!}{
\begin{tabular}{llllllllccc}
\toprule
\multirow{4}{*}{\textbf{Model}} 
& \multirow{4}{*}{\textbf{Method}} 
& \multicolumn{6}{c}{\textbf{ToolBench}} 
& \multicolumn{3}{c}{\textbf{MCP-Universe}} \\
\cmidrule(lr){3-8}
\cmidrule(lr){9-11}
& 
& \multicolumn{2}{c}{\textbf{G1}} 
& \multicolumn{2}{c}{\textbf{G2}} 
& \multicolumn{2}{c}{\textbf{G3}}
& \multirow{2}{*}{\textbf{Location Navigation}}
& \multirow{2}{*}{\textbf{Repository Management}}
& \multirow{2}{*}{\textbf{Financial Analysis}} \\
\cmidrule(lr){3-4}
\cmidrule(lr){5-6}
\cmidrule(lr){7-8}
& & Pass. & Win. & Pass. & Win. & Pass. & Win. &  &  &  \\
\midrule

\multirow{6}{*}{Qwen-3-235B}
& ReACT        
    &  50.5 &  --   
    &  53.5 &  --   
    &  46.0 &  --   
    &  11.10 &  9.09  &  50.0 \\
& DFSDT        
    &  57.0 &  53.8 
    &  61.5 &  67.5 
    &  48.8 &  56.8 
    &  11.10 &  12.12 &  50.0 \\
& LATS          
    &  62.5 &  59.3 
    &  78.0 &  70.3 
    &  77.8 &  83.3 
    &  15.54 &  15.15 &  52.5 \\
& ToolChain* &  65.0 &  62.8 
    &  79.3 &  72.5 
    &  78.0 &  83.5 
    &  16.65 &  18.18 &  55.0 \\
& Tool-Planner 
    &  60.3 &  58.0 
    &  70.5 &  68.8 
    &  65.5 &  72.3 
    &  13.32 &  12.12 &  52.5 \\
& \cellcolor{Red}ToolGate 
    & \cellcolor{Red}\textbf{68.3} & \cellcolor{Red}\textbf{65.5} 
    & \cellcolor{Red}\textbf{82.5} & \cellcolor{Red}\textbf{78.0} 
    & \cellcolor{Red}\textbf{81.0} & \cellcolor{Red}\textbf{82.3} 
    & \cellcolor{Red}\textbf{18.87}
    & \cellcolor{Red}\textbf{21.21}
    & \cellcolor{Red}\textbf{60.0} \\
\midrule

\multirow{6}{*}{Deepseek V3.2}
& ReACT        
    &  52.0 &  48.5   
    &  55.3 &  51.0   
    &  48.5 &  53.5   
    &  12.21 &  12.12 &  52.5 \\
& DFSDT        
    &  58.5 &  55.0 
    &  63.0 &  69.3 
    &  50.3 &  58.8 
    &  13.32 &  15.15 &  55.0 \\
& LATS          
    &  65.3 &  61.8 
    &  80.0 &  72.5 
    &  80.3 &  85.5 
    &  17.76 &  18.18 &  60.0 \\
& ToolChain* &  68.8 &  65.0 
    &  82.5 &  75.3 
    &  81.0 &  88.8 
    &  18.87 &  21.21 &  62.5 \\
& Tool-Planner 
    &  62.5 &  60.3 
    &  73.8 &  70.0 
    &  68.0 &  75.5 
    &  15.54 &  15.15 &  57.5 \\
& \cellcolor{Red}ToolGate 
    & \cellcolor{Red}\textbf{72.0} & \cellcolor{Red}\textbf{70.3} 
    & \cellcolor{Red}\textbf{85.5} & \cellcolor{Red}\textbf{80.0} 
    & \cellcolor{Red}\textbf{85.3} & \cellcolor{Red}\textbf{81.3} 
    & \cellcolor{Red}\textbf{22.20}
    & \cellcolor{Red}\textbf{24.24}
    & \cellcolor{Red}\textbf{67.5} \\

\midrule

\multirow{6}{*}{GPT-5.2}
& ReACT        
    &  63.5 &  62.8   
    &  65.0 &  63.2   
    &  58.3 &  59.5   
    &  18.87 &  24.24 &  65.0 \\
& DFSDT        
    &  70.0 &  68.5 
    &  75.3 &  78.0 
    &  63.8 &  70.5 
    &  19.98 &  27.27 &  67.5 \\
& LATS          
    &  80.3 &  78.8 
    &  88.5 &  85.3 
    &  85.0 &  90.8 
    &  28.86 &  36.36 &  82.5 \\
& ToolChain* &  82.8 &  80.0 
    &  90.5 &  88.3 
    &  88.5 &  92.5 
    &  29.97 &  39.39 &  85.0 \\
& Tool-Planner 
    &  75.5 &  72.3 
    &  82.0 &  80.5 
    &  78.3 &  85.0 
    &  25.53 &  30.30 &  75.0 \\
& \cellcolor{Red}ToolGate 
    & \cellcolor{Red}\textbf{85.5} & \cellcolor{Red}\textbf{83.5} 
    & \cellcolor{Red}\textbf{93.0} & \cellcolor{Red}\textbf{90.5} 
    & \cellcolor{Red}\textbf{91.8} & \cellcolor{Red}\textbf{95.3} 
    & \cellcolor{Red}\textbf{35.52}
    & \cellcolor{Red}\textbf{45.45}
    & \cellcolor{Red}\textbf{90.0} \\
\midrule

\multirow{6}{*}{Gemini 3 Pro}
& ReACT        
    &  60.0 &  60.5   
    &  63.8 &  57.0   
    &  55.5 &  56.5   
    &  16.65 &  21.21 &  62.5 \\
& DFSDT        
    &  68.3 &  65.5 
    &  72.0 &  75.8 
    &  60.3 &  68.5 
    &  17.76 &  24.24 &  65.0 \\
& LATS          
    &  78.5 &  75.3 
    &  85.8 &  82.0 
    &  82.5 &  88.3 
    &  26.64 &  33.33 &  77.5 \\
& ToolChain* &  80.0 &  78.5 
    &  88.3 &  85.0 
    &  85.8 &  90.0 
    &  27.75 &  36.36 &  80.0 \\
& Tool-Planner 
    &  73.8 &  70.0 
    &  80.5 &  78.3 
    &  75.0 &  82.8 
    &  22.20 &  27.27 &  72.5 \\
& \cellcolor{Red}ToolGate 
    & \cellcolor{Red}\textbf{83.0} & \cellcolor{Red}\textbf{80.5} 
    & \cellcolor{Red}\textbf{91.3} & \cellcolor{Red}\textbf{88.0} 
    & \cellcolor{Red}\textbf{90.0} & \cellcolor{Red}\textbf{93.5} 
    & \cellcolor{Red}\textbf{33.30}
    & \cellcolor{Red}\textbf{42.42}
    & \cellcolor{Red}\textbf{87.5} \\

\bottomrule
\end{tabular}
}
\end{table*}

\subsection{Evaluation Metrics}
For ToolBench, we adopt two evaluation metrics from ToolEval \citep{qintoolllm}. The first metric is \textbf{Pass Rate}, computed as the proportion of successfully completed tasks, which reflects overall task-solving capability. The second metric is \textbf{Win Rate}, where we compare the execution plans and results produced by our framework with those generated by Qwen-3 235B-ReACT and request LLMs judges to determine which solution is superior. If our method yields a better solution, we mark it as a win; if it is equivalent or worse, we mark it as a tie or loss. Win Rate therefore measures both reasoning quality and execution superiority.

For MCP-Universe, we evaluate \textbf{Success Rate} and execution stability. Many tasks in MCP-Universe involve relatively fewer tool invocation steps but arise from real-world complex systems.

\subsection{Baselines}
We compare our framework against the following representative tool-use and planning baselines:
 \textbf{ReACT} \citep{yao2022react}, 
\textbf{DFSDT} \citep{qintoolllm},
 \textbf{LATS} \cite{zhou2024language}, 
 \textbf{ToolChain*} \citep{zhuangtoolchain},
 \textbf{Tool-Planner} \citep{liu2025toolplanner}, More baselise details are provided in Appendix~\ref{sec:baselines}.

\subsection{Models}
We evaluate our framework across a range of large language models to verify generality and robustness. Proprietary models include \textbf{Gemini 3 Pro} \cite{Gemini3}, \textbf{GPT-5.2} \cite{GPT5}. Open-source models include \textbf{DeepSeek V3.2} \cite{liu2025deepseek}, \textbf{Qwen3-235B-A22B-Instruct-2507} \cite{yang2025qwen3}. These models cover heterogeneous training paradigms, reasoning capabilities, and scales. While we use \textbf{Qwen3-embedding-0.6B} and \textbf{Qwen3-Reranker-0.6B} \cite{zhang2025qwen3} for tool embedding and retrieval.

\section{Experiments}

\subsection{Main Results}

As shown in Table~\ref{tab:toolgate_main_result}, we conduct comprehensive evaluations on ToolBench (G1/G2/G3) and MCP-Universe. 

\textbf{For ToolBench.} The results show that ToolGate achieves the best or near-best performance across all models and all evaluation benchmarks. On ToolBench, ToolGate leads to substantial improvements in both Pass Rate and Win Rate across all three task groups. For instance, under GPT-5.2, ToolGate reaches \textbf{85.5 / 83.5}, \textbf{93.0 / 90.5}, and \textbf{91.8 / 95.3} on G1/G2/G3 respectively, outperforming the strongest baseline ToolChain* by approximately $4-6\%$ in Win Rate. Similar improvements are consistently observed on Qwen-3-235B, DeepSeek V3.2, and Gemini 3 Pro, demonstrating that ToolGate is \textbf{model-agnostic} and provides stable enhancement to tool reasoning and execution capabilities across different LLM backbones.

\begin{table*}[t]
\centering
\caption{Comprehensive ablation study of the Hoare logic verification module. We compare the full ToolGate architecture against variants: \textbf{No $\{P\}$ check} (skips pre-condition validation) and \textbf{No $\{Q\}$ check} (skips post-condition assertion). \textbf{MCP-Avg} represents the mean success rate of MCP subtasks.}
\label{tab:hoare_ablation_final}
\resizebox{\textwidth}{!}{
\begin{tabular}{llcccccccccc}
\toprule
\multirow{3}{*}{\textbf{Model}} & \multirow{3}{*}{\textbf{Method}} & \multicolumn{6}{c}{\textbf{ToolBench}} & \multicolumn{4}{c}{\textbf{MCP-Universe}} \\
\cmidrule(lr){3-8} \cmidrule(lr){9-12}
& & \multicolumn{2}{c}{\textbf{G1}} & \multicolumn{2}{c}{\textbf{G2}} & \multicolumn{2}{c}{\textbf{G3}} & \multirow{2}{*}{\textbf{Loc.}} & \multirow{2}{*}{\textbf{Repo.}} & \multirow{2}{*}{\textbf{Fin.}} & \multirow{2}{*}{\textbf{MCP-Avg}} \\
\cmidrule(lr){3-4} \cmidrule(lr){5-6} \cmidrule(lr){7-8}
& & Pass. & Win. & Pass. & Win. & Pass. & Win. & & & & \\
\midrule
\multirow{7}{*}{DeepSeek V3.2} 
& ReACT & 52.0 & 48.5 & 55.3 & 51.0 & 48.5 & 53.5 & 12.2 & 12.1 & 52.5 & 25.6 \\
& DFSDT & 58.5 & 55.0 & 63.0 & 69.3 & 50.3 & 58.8 & 13.3 & 15.2 & 55.0 & 27.8 \\
& ToolChain* & 68.8 & 65.0 & 82.5 & 75.3 & 81.0 & 88.8 & 18.9 & 21.2 & 62.5 & 34.2 \\
\cmidrule(lr){2-12}
& ToolGate w/o Hoare & 57.2 & 53.8 & 61.5 & 67.5 & 49.8 & 57.5 & 12.8 & 14.5 & 54.2 & 27.2 \\
& \textbf{-- No $\{P\}$ check} & 67.8 & 66.5 & 79.5 & 77.2 & 78.4 & 82.8 & 19.8 & 21.5 & 62.2 & 34.5 \\
& \textbf{-- No $\{Q\}$ check} & 63.2 & 61.0 & 71.5 & 72.8 & 70.8 & 73.5 & 16.2 & 18.2 & 58.2 & 30.9 \\
& \textbf{ToolGate (Full)} & \textbf{72.0} & \textbf{70.3} & \textbf{85.5} & \textbf{80.0} & \textbf{85.3} & \textbf{81.3} & \textbf{22.2} & \textbf{24.2} & \textbf{67.5} & \textbf{38.0} \\
\midrule
\multirow{7}{*}{GPT-5.2} 
& ReACT & 63.5 & 62.8 & 65.0 & 63.2 & 58.3 & 59.5 & 18.9 & 24.2 & 65.0 & 36.0 \\
& DFSDT & 70.0 & 68.5 & 75.3 & 78.0 & 63.8 & 70.5 & 20.0 & 27.3 & 67.5 & 38.3 \\
& ToolChain* & 82.8 & 80.0 & 90.5 & 88.3 & 88.5 & 92.5 & 30.0 & 39.4 & 85.0 & 51.5 \\
\cmidrule(lr){2-12}
& ToolGate w/o Hoare & 69.2 & 67.5 & 74.5 & 76.8 & 62.5 & 69.2 & 19.5 & 26.8 & 66.5 & 37.6 \\
& \textbf{-- No $\{P\}$ check} & 81.2 & 79.5 & 89.2 & 88.0 & 86.4 & 90.5 & 31.0 & 41.5 & 85.0 & 52.5 \\
& \textbf{-- No $\{Q\}$ check} & 75.5 & 74.0 & 82.8 & 83.5 & 79.2 & 82.0 & 25.5 & 33.5 & 79.6 & 46.2 \\
& \textbf{ToolGate (Full)} & \textbf{85.5} & \textbf{83.5} & \textbf{93.0} & \textbf{90.5} & \textbf{91.8} & \textbf{95.3} & \textbf{35.5} & \textbf{45.5} & \textbf{90.0} & \textbf{57.0} \\
\bottomrule
\end{tabular}
}
\end{table*}

\textbf{For MCP-Universe.} The advantage of ToolGate becomes even more pronounced, which emphasizes long-horizon tool dependencies and real-world execution robustness. ToolGate yields $3-7\%$  improvements over ToolChain* in \textit{Location Navigation} and \textit{Repository Management}, and delivers state-of-the-art performance on \textit{Financial Analysis}. Notably, GPT-5.2 with ToolGate achieves \textbf{45.45} in Repository Management and \textbf{90.0} in Financial Analysis, substantially surpassing all competing systems. These results suggest that ToolGate not only improves task success rates, but also significantly enhances stability and robustness when executing complex tool chains.

\textbf{On Multi-tool instructions tasks.} Experimental results indicates that these gains are not merely due to stronger heuristics or more aggressive exploration, but primarily arise from ToolGate’s Hoare-logic-based formal constraint mechanism. During reasoning, the system explicitly maintains a trusted state set $S$ and constructs a Hoare Triple $\{P\}\ C\ \{Q\}$ for each tool invocation, enforcing precondition and postcondition validation. This enables ToolGate to significantly reduce error accumulation in complex ToolBench tasks such as G2 and G3, resulting in higher and more stable Win Rates; meanwhile, in MCP-Universe, it effectively mitigates long-horizon reasoning drift, leading to sustained performance gains under multi-tool dependency and real-world execution constraints.

\subsection{Ablation Studies}

To evaluate the structural dependency of ToolGate on its formal verification mechanism, we conducted a systematic ablation study across both DeepSeek V3.2 and GPT-5.2. We specifically isolated the Hoare logic module to observe its impact on tool-use efficacy. As detailed in Table \ref{tab:hoare_ablation_final}, the results reveal a critical finding: removing the formal verification layer leads to a performance level that falls marginally below the standard DFSDT baseline.

For instance, with GPT-5.2, the MCP-Avg success rate for the ToolGate without Hoare filtering is 37.6\%, which is slightly lower than the 38.3\% achieved by DFSDT. This trend is consistent across DeepSeek V3.2 as well. This indicates that without the pruning capabilities provided by Hoare logic, the underlying search architecture of ToolGate becomes less efficient than a conventional depth-first search strategy.

The results demonstrate that Hoare logic verification is the key factor behind ToolGate's search efficiency. The two fundamental components of the Hoare logic framework, the precondition $\{P\}$ and the postcondition $\{Q\}$ serve complementary functions in guiding tool-invocation decisions. Empirical evidence highlights that the absence of $\{Q\}$ checks is substantially more detrimental than the absence of $\{P\}$ checks. For instance, on GPT-5.2, the MCP-Avg success rate drops by 10.8\% when $\{Q\}$ checks are removed, compared to a 4.5\% decrease when $\{P\}$ checks are omitted. The full version of ToolGate enforces a rigorous $\{P\} C \{Q\}$ logical closed-loop, ensuring that every step within the search process is both formally valid and substantively effective. the performance gap between the full ToolGate model and its ablated counterpart confirms that formal logic is the primary catalyst for superior reliability and task success.

\begin{figure}
    \centering
    \includegraphics[width=1\linewidth]{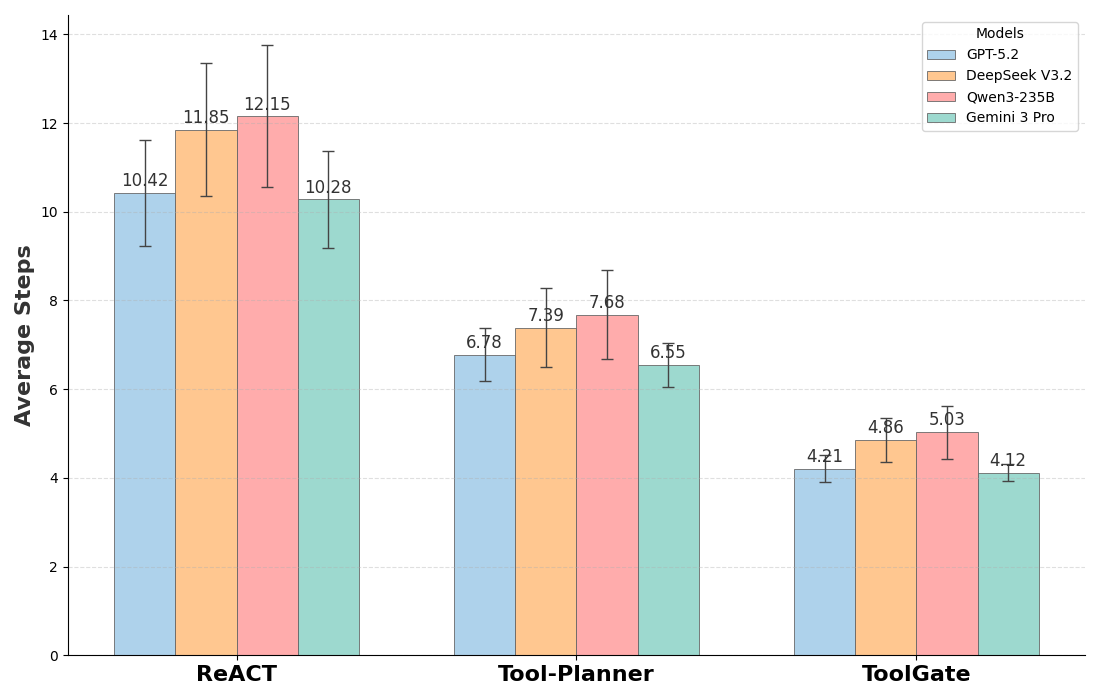}
    \caption{Comparison of Average Tool-Calling Steps in Tool-Bench.}
    \label{fig:calling_steps_comparison}
\end{figure}


\subsection{Tool Reasoning Efficiency}

To evaluate the search efficiency of ToolGate, we focus on the average number of tool-calling steps required to complete tasks. This metric serves as a proxy for the model's ability to navigate complex state-spaces without redundant exploration.

As shown in Figure \ref{fig:calling_steps_comparison}, ToolGate consistently achieves the most concise tool-calling trajectories across both GPT-5.2 and DeepSeek V3.2 backbones. Specifically, when using GPT-5.2, ToolGate reduces the average calling steps from $6.78$ to $4.21$, representing a $37.9\%$ improvement in efficiency. While traditional methods like ReACT and Tool-Planner often fall into "trial-and-error" loops due to a lack of environmental awareness, ToolGate maintains a trajectory close to the theoretical optimal path.

The efficiency of ToolGate is primarily attributed to the Hoare logic verification module. In the vast state-space of Tool-Bench, logical conflicts between tool preconditions and environmental states are frequent. Unlike Tool-Planner, which explores branches based on probabilistic heuristics, ToolGate applies formal constraints to prune the search tree. By verifying the feasibility of a tool call before execution, the system effectively collapses the search space, eliminating branches that are logically destined to fail.

\subsection{Fine-grained Rejection Distribution}

To further investigate the internal decision-making mechanism of ToolGate, we conducted a comprehensive trace of all tool-invocation attempts during the evaluation. Our results indicate that in high-complexity benchmarks such as \textbf{MCP-Universe}, the formal verification layer intercepts approximately \textbf{29.4\%} of the total tool-calling requests. Based on a fine-grained analysis of these rejections, we categorize the findings into three key areas:

\textbf{Static Pruning via $\{P\}$} The pre-condition check primarily filters \textbf{parametric hallucinations} and \textbf{state dependency violations}. By intercepting invalid IDs and out-of-sequence calls before execution, $\{P\}$ significantly reduces computational overhead and prevents the search tree from expanding into invalid branches.

\textbf{Dynamic Rectification via $\{Q\}$ }. The post-condition assertion captures sophisticated failures that standard models miss, By mandating semantic alignment and state consistency, $\{Q\}$ identifies logically vacuous steps and triggers immediate backtracking, preventing cascading errors.

While $\{P\}$ optimizes \textbf{efficiency} by pruning 17.6\% of invalid paths statically, $\{Q\}$ ensures \textbf{task success} by dynamically rectifying the remaining 11.8\% of logical drifts. Together, they form a logical closed-loop that anchors the agent to the correct semantic trajectory. 

\begin{table}[t]
\centering
\caption{Fine-grained analysis of logical rejections across Hoare components. Rates are normalized against the total number of tool invocations in the MCP-Universe benchmark.}
\label{tab:rejection_breakdown_final}
\resizebox{\columnwidth}{!}{
\begin{tabular}{llc}
\toprule
\textbf{Verification Phase} & \textbf{Specific Error Sub-category} & \textbf{Abs. Rate (\%)} \\
\midrule
\multirow{3}{*}{Pre-condition $\{P\}$} & Value/Entity Hallucination & 8.4\% \\
& Schema \& Format Violation & 5.1\% \\
& State Dependency Missing & 4.1\% \\
\cmidrule(lr){2-3}
& \textit{Subtotal $\{P\}$ Rejections} & \textbf{17.6\%} \\
\midrule
\multirow{3}{*}{Post-condition $\{Q\}$} & Empty/Null& 6.3\% \\
& Semantic Constraint Mismatch & 3.7\% \\
& State Update Inconsistency & 1.8\% \\
\cmidrule(lr){2-3}
& \textit{Subtotal $\{Q\}$ Rejections} & \textbf{11.8\%} \\
\midrule
\textbf{Total} & \textbf{Combined Rejection Rate} & \textbf{29.4\%} \\
\bottomrule
\end{tabular}
}
\end{table}

\section{Conclusions}
In this paper, we introduces ToolGate, a comprehensive method to evaluate the critical decision-making and gatekeeping capabilities of Large Language Models (LLMs) in tool-use scenarios. By shifting the evaluation focus from mere execution success to the nuanced assessment of when to invoke or refuse a tool, ToolGate reveals a prevalent tendency toward over-reliance in current state-of-the-art models, particularly when they encounter ambiguous, unauthorized, or high-risk instructions. These findings underscore the urgent necessity of balancing functional proficiency with robust decision-making frameworks. Ultimately, ToolGate provides both a diagnostic tool and a foundational framework for the development of safer, more reliable, and more autonomous AI agents in real-world applications.

\section*{Limitations}

Despite its contributions, several limitations of ToolGate should be acknowledged. First, while the benchmark covers a diverse range of scenarios, its current scope is primarily restricted to text-based and structured data interactions, leaving multi-modal tools and long-chain, multi-step collaborative tasks as areas for future expansion. Second, the evaluation environment is largely static, which may not fully capture the complexities of real-world API dynamics, such as network latency, rate limits, or fluctuating data states that can interfere with real-time decision-making. Furthermore, our evaluation metrics remain predominantly quantitative; future work is needed to develop more fine-grained qualitative assessments of a model's explanatory reasoning and its ability to proactively solicit missing information from users. Finally, the potential for prompt-based bias remains, as strategies optimized for specific models may not generalize perfectly across the entire landscape of open-source LLMs.

\section*{Ethics Considerations}

ToolGate is developed as a general framework for formal, verifiable, and responsible tool use in large language model reasoning. All experiments are conducted on publicly available benchmarks or open-source tool environments, and no private or personally identifiable information is collected, accessed, or utilized throughout our work. The tools invoked in our experiments are either simulated environments or publicly documented APIs with appropriate usage permissions.

Our framework does not generate, store, or infer sensitive personal attributes, nor does it target any specific demographic groups. Instead, ToolGate focuses on improving reliability, interpretability, and safety in model-based tool invocation by enforcing logical constraints and verifiable execution conditions. During evaluation, we strictly follow the licenses and terms of use associated with the released LLMs, datasets, APIs, and benchmark platforms.

Furthermore, we emphasize that ToolGate is designed to enhance trustworthy AI behaviors rather than to bypass safeguards or enable harmful automation. The methodology can be integrated with additional safety filters, auditing processes, and access control mechanisms when deployed in real-world systems. We believe this contributes positively toward building transparent, controllable, and ethically aligned AI tool-use systems.

\bibliography{anthology,custom}
\bibliographystyle{acl_natbib}

\appendix

\section{Datasets Detailed}
\label{sec:appendixA}
\subsection{ToolBench Dataset}

\textbf{ToolBench} \cite{qintoolllm} is a large-scale instruction tuning and evaluation dataset proposed in the ToolLLM framework, aiming to systematically assess and enhance large language models’ ability in tool selection, parameter planning, and executable API invocation in real-world environments. During construction, ToolBench first collects \textbf{16,464 real RESTful APIs} from RapidAPI Hub, covering \textbf{49 functional categories} (e.g., weather, social media, e-commerce, and mapping services), and extracts structured metadata including API names, documentation, parameter schemas, and usage examples. Based on these APIs, natural-language task instructions are automatically generated using LLMs, and a depth-first search based decision tree (DFSDT) is employed to discover feasible \textbf{tool-call trajectories} as solution paths.

In terms of scale, ToolBench provides more than \textbf{126K instruction--solution path pairs} under large API spaces, with multiple train/test splits designed to test generalization under unseen instructions, unseen tools, and even unseen tool categories. In addition, ToolBench includes a ``classic task set'' covering \textbf{8 representative tool environments} such as OpenWeather, VirtualHome, and WebShop, each containing about 100 manually verified task instructions and 7--15 tool interfaces, enabling more fine-grained ablation and comparative studies.

For evaluation, ToolBench integrates the ToolEval framework to conduct execution-level assessment on generated API call sequences. Typical metrics include \textbf{Pass Rate} (task completion), \textbf{Win Rate}. Some works further adopt ``Plan.EM'' and ``Act.EM'' to decouple planning quality and execution quality. Due to its realistic and large-scale API space, ToolBench has become a widely adopted benchmark and data source for a series of subsequent tool-use research works.

\vspace{6pt}
\subsection{MCP-Universe Benchmark}

\textbf{MCP-Universe} \cite{luo2025mcpuniverse} is a comprehensive benchmark proposed for evaluating large language models under the \textbf{Model Context Protocol (MCP)} paradigm, focusing on their capability to perform complex tasks via interaction with \emph{real} MCP servers. Unlike offline tool-use datasets, MCP-Universe directly connects to real running MCP services, emphasizing long-horizon interaction, unknown tool discovery, and robust execution under dynamic environments.

MCP-Universe spans \textbf{6 core task domains} and \textbf{11 different MCP servers}, including Location Navigation, Repository Management, Financial Analysis, 3D Design, Browser Automation, and Web Searching. The benchmark contains \textbf{231 task instances} in total, with multiple benchmark configurations derived from different combinations of environments and tools. In addition, the benchmark defines \textbf{84 unique evaluators} to cover different evaluation dimensions such as structural correctness, logical soundness, and consistency with dynamic data sources.

In terms of task distribution, the benchmark is designed to be representative while maintaining reasonable balance: Web Search tasks account for approximately 23.8\% (55 tasks), Location Navigation 19.5\% (45 tasks), Financial Analysis 17.3\% (40 tasks), Browser Automation 16.9\% (39 tasks), Repository Management 14.3\% (33 tasks), and 3D Design 8.2\% (19 tasks). Tasks generally require agents to interact with multiple MCP tools across several rounds, performing complex objectives such as route planning, repository manipulation, portfolio analysis, or automated browser operations.

MCP-Universe further distinguishes \textbf{three categories of execution-level evaluators}:  

(1) \textbf{Format Evaluators}, checking whether model outputs follow the MCP calling specification;  

(2) \textbf{Static Evaluators}, validating correctness for time-invariant tasks;  

(3) \textbf{Dynamic Evaluators}, querying real-time data sources to construct ground-truth for time-sensitive tasks such as financial prices or navigation.

\section{Baselines Detailed}
\label{sec:baselines}
Our method is compared against several state-of-the-art and comprehensive baselines, covering the following benchmark settings:

 \textbf{ReACT} \citep{yao2022react}, which alternates reasoning Thought and execution Action, forming a linear interaction process between language reasoning and tool invocation. It is one of the most widely used baselines for tool-augmented LLMs.
 
\textbf{DFSDT} \citep{qintoolllm}, which adopts a depth-first search mechanism to explore reasoning–tool trajectories. Whenever the model reaches an erroneous path, DFSDT exposes the full failure history back to the model, enabling re-planning and maximizing exploration space.

 \textbf{LATS} \cite{zhou2024language}, which leverages look-ahead tree search to expand multiple candidate tool sequences and evaluates their expected effectiveness, demonstrating strong planning ability in complex multi-step scenarios.
 
 \textbf{ToolChain*} \citep{zhuangtoolchain}, which explicitly constructs a tool chain to model multi-step dependencies and guides LLMs to complete sequential tool execution. Although it enhances structured reasoning for multi-tool tasks, its effectiveness still primarily relies on LLM-based natural language reasoning rather than formal execution constraints.
 
 \textbf{Tool-Planner} \citep{liu2025toolplanner}, which incorporates explicit external planning modules to control tool sequence generation, combining retrieval, candidate filtering, and structured planning strategies to improve global execution coherence and decision reliability.

 \section{Environments}

All experiments are conducted under a unified tool execution environment. ToolBench APIs are treated as callable functional nodes, while MCP-Universe tools are executed in an official sandbox with real execution feedback, including realistic execution latency, tool failure signals, and state-dependent output variations. All models are accessed through their official APIs, with the decoding temperature fixed to 0.2 to minimize randomness in reasoning and tool planning behavior.  For each benchmark, all systems share identical task instructions, tool descriptions, execution limits, and termination conditions.

For the retrieval module, we construct tool semantic representations using an embedding-based retrieval framework. Specifically, we adopt the Qwen3-embedding-0.6b \cite{zhang2025qwen3} model to encode tool descriptions, functional semantics, argument specifications, and usage documentation into dense vectors. During reasoning, the intermediate tool requirement representation is encoded in the same embedding space, and the Top-$K$ candidate tools are retrieved using cosine similarity. We set $K = 10$ by default unless otherwise specified. Following retrieval, a reranking model is applied to improve tool selection accuracy within the narrowed candidate set. We employ a lightweight LLM-based reranker built upon Qwen3-Reranker-0.6B \cite{zhang2025qwen3}, which jointly considers the current reasoning context, symbolic state, and candidate tool semantics to estimate contextual suitability.

\section{Detailed Rejection Analysis}

\subsection{Pre-condition $\{P\}$ Validation} 

The pre-condition check accounts for \textbf{17.6\%} of all rejections, functioning as a "static firewall" that blocks invalid actions before they are executed.
\begin{itemize}
    \item \textbf{Parametric Hallucination (8.4\%):} This is the most prevalent error type. When facing vast toolsets, LLMs often generate parameters based on intuition—such as hallucinating file IDs or directory paths—rather than grounded retrieval. By enforcing symbolic link validation, $\{P\}$ intercepts these requests before execution, significantly reducing computational overhead and token consumption.
    \item \textbf{State Dependency Violation (4.1\%):} Models occasionally bypass necessary operational sequences, such as attempting to modify a file without first obtaining the required permissions or handles. $\{P\}$ enforces strict logical and temporal constraints, ensuring that every invocation is predicated on a valid environmental state.
\end{itemize}

\subsection{Post-condition $\{Q\}$ Assertion} 

Although post-execution assertions trigger less frequently (\textbf{11.8\%}), they address sophisticated logical failures that conventional search models, such as DFSDT, typically fail to detect.
\begin{itemize}
    \item \textbf{Silent Failures (6.3\%):} In complex tasks, APIs often return a successful status code (e.g., HTTP 200) despite providing an empty or vacuous response (e.g., \texttt{results: []}). Without $\{Q\}$ verification, an agent might interpret this as successful progress and continue down a futile search path. The $\{Q\}$ assertion mandates a \textit{non-empty} result check, identifying these "logical voids" and triggering an immediate backtrack.
    \item \textbf{Semantic and State Alignment (5.5\%):} This includes mismatches in semantic constraints (3.7\%) and inconsistencies in state updates (1.8\%). This confirms that $\{Q\}$ can capture subtle deviations between tool outputs and user intent, ensuring the search trajectory remains anchored to the correct semantic path.
\end{itemize}

\subsection{Synergetic Effects and Logical Closure}
Our analysis reveals that $\{P\}$ and $\{Q\}$ constitute a robust logical closed-loop. The high rejection rate of $\{P\}$ (\textbf{17.6\%}) primarily improves \textbf{search efficiency} by pruning obvious error branches to save tokens and time. Conversely, the precision-driven interceptions of $\{Q\}$ (\textbf{11.8\%}) are the primary determinants of \textbf{task success}. By identifying technically successful but logically flawed steps, $\{Q\}$ prevents the accumulation of cascading errors as a major bottleneck in unverified agentic systems.

\subsection{Prompt Template}

\onecolumn{
\begin{mybox}{System Prompt for LLM Reasoning}
    \begin{verbatim}
You are a helpful assistant that can use tools to answer user questions.

You have access to a set of tools and a symbolic state that tracks verified facts.

**Important Control Tokens:**
- When you need to use a tool, output: `<start_call_tool>`
- After tool results are provided, they will be wrapped in: 
  `<start_tool_result>...</end_tool_result>`
- When you finish using tools, output: `<end_call_tool>`

**State Information:**
The current symbolic state contains verified facts. Use this information to:
1. Check if you have enough information to answer directly
2. Determine what information is missing and needs to be retrieved via tools
3. Understand what tools can be called based on the current state

**Tool Calling Process:**
1. Think about what information you need
2. Output `<start_call_tool>` followed by a brief description of what you need
3. Wait for tool results
4. Continue reasoning with the new information
5. Repeat if needed, or provide the final answer

**CRITICAL: When Tools Fail - Keep Trying!**
- If a tool call fails, DO NOT give up immediately. Consider:
  * Try a different tool that might provide similar information
  * Try the same tool with different parameters
  * Try alternative approaches or search strategies
  * Think about what other information sources might help
- Only provide a final answer when you are CONFIDENT you have:
  * Successfully retrieved the necessary information, OR
  * Exhausted all reasonable tool options and can provide a helpful answer 
    based on available information
- Do NOT end reasoning prematurely just because one tool failed
- Be persistent and creative in finding alternative solutions

**Output Format:**
- If you can answer directly: Provide the answer without `<start_call_tool>`
- If you need tools: Output `<start_call_tool>` followed by your reasoning 
  about what tool to use
- If tools fail: Think about alternatives and try again with 
  `<start_call_tool>`
    \end{verbatim}
\end{mybox}
}

\onecolumn{
\begin{mybox}{Example User Message Format}
    \begin{verbatim}
**Current State:**
query: Find me a tutorial video about machine learning on YouTube
topic: machine learning
platform: YouTube
content_type: tutorial video

**Tool Results:**
(Empty on first call)

**User Query:** Find me a tutorial video about machine learning on YouTube

**Your Task:**
Think step by step. If you need to use tools, output `<start_call_tool>` 
followed by a description of what you need.
If a tool fails, think about alternative approaches and try other tools 
before giving up.
Only provide the final answer when you are CONFIDENT you have enough 
information or have exhausted all reasonable options.
    \end{verbatim}
\end{mybox}
}

\section{Planning Process}

Algorithm \ref{alg:forward_execution} formalizes our Forward Execution framework, which integrates Contract Verification into the LLM's reasoning loop. The procedure begins by initializing the environment state $S_0$ from the user query and context. At each step $k$, the LLM acts as a controller, deciding whether to conclude the task with an answer or invoke an external tool.

\onecolumn{
\begin{algorithm}[h]
\caption{Forward Execution with Contract Verification}
\label{alg:forward_execution}
\begin{algorithmic}[1]
\Require Query $q$, Context $H$, Max iterations $K_{\max}$, Toolset $\mathcal{T}$
\Ensure Final answer $a$ or failure signal
\State $S_0 \gets \text{InitState}(q, H)$, $\mathcal{T}_{\text{failed}} \gets \emptyset$
\For{$k = 0$ \textbf{to} $K_{\max}-1$} \label{line:loop}
    \State $\text{act}_k \gets \text{LLM\_Reason}(q, H, \text{summary}(S_k))$ \Comment{Generate reasoning step}
    
    \If{$\text{act}_k$ is $\text{Answer}(a)$} 
        \State \textbf{return} $a$ \Comment{Task successfully completed}
    \EndIf

    \If{$\text{act}_k$ is $\text{CallTool}(\textit{desc})$}
        \State $\mathcal{T}_{\text{cand}} \gets \text{Rerank}(\text{Retrieve}(q, S_k))$ \Comment{Top-$N$ candidate tools}
        \State $\text{is\_updated} \gets \text{False}$
        
        \For{$t \in \mathcal{T}_{\text{cand}} \setminus \mathcal{T}_{\text{failed}}$} \Comment{Iterate through valid candidates}
            \If{$S_k \not\models \phi_{pre}(t)$} \label{line:pre} \Comment{Check Precondition Contract}
                \State \textbf{continue} 
            \EndIf
            
            \State $\theta_t \gets \text{GenParams}(t, S_k)$ \Comment{Synthesize arguments}
            \State $r_t \gets \text{Execute}(t, \theta_t)$ \Comment{Tool invocation}
            
            \If{$\text{Verify}(S_k, r_t, \phi_{post}(t))$} \label{line:post} \Comment{Verify Postcondition Contract}
                \State $S_{k+1} \gets \text{Update}(S_k, r_t)$ \Comment{Commit to state transition}
                \State $\text{is\_updated} \gets \text{True}$
                \State \textbf{break} 
            \Else
                \State $\mathcal{T}_{\text{failed}} \gets \mathcal{T}_{\text{failed}} \cup \{t\}$ \Comment{Mark tool as unreliable}
            \EndIf
        \EndFor
        
        \If{not $\text{is\_updated}$} 
            \State \textbf{return} \text{Fail} \Comment{No viable tools satisfy contracts}
        \EndIf
    \EndIf
\EndFor
\State \textbf{return} \text{Timeout}
\end{algorithmic}
\end{algorithm}
}

\section{Examples}


\onecolumn{
\begin{mybox}{Tool Calling Example I}
    \begin{itemize}
        \item \textbf{Query:} ``Find me a tutorial video about machine learning on YouTube''
        
        \item \textbf{Step 0: Initial State}
        \begin{verbatim}
S_0 = {
    "query": "Find me a tutorial video about machine learning on YouTube",
    "topic": "machine learning",
    "platform": "YouTube",
    "content_type": "tutorial video"
}
        \end{verbatim}
        
        \item \textbf{Step 1: LLM Reasoning}
        \begin{verbatim}
Conversation History: []
LLM Response:
"I need to search for machine learning tutorial videos on YouTube. 
Let me use a YouTube search tool to find relevant videos.

<start_call_tool>
I need to search for machine learning tutorial videos on YouTube
<end_call_tool>"

Action: start_call_tool
        \end{verbatim}
        
        \item \textbf{Step 1: Tool Retrieval}
        \begin{verbatim}
Retrieved candidates (top-5 after reranking):
1. Simple_YouTube_Search_Search (score: 0.92)
2. YouTube_Video_Info (score: 0.85)
3. YouTube_Channel_Search (score: 0.78)
4. Video_Search_API (score: 0.72)
5. Content_Search (score: 0.68)
        \end{verbatim}
        
        \item \textbf{Step 1: Precondition Checking}
        \begin{verbatim}
Checking Simple_YouTube_Search_Search:
  Precondition: exists(query)
  State S_0: {"query": "Find me a tutorial video...", ...}
  Check: exists(query) = true 
        \end{verbatim}
        
        \item \textbf{Step 1: Tool Execution}
        \begin{verbatim}
Tool: Simple_YouTube_Search_Search
Params: {"query": "machine learning tutorial"}
Tool Config: {
    "category": "Data",
    "api_name": "Search",
    "tool_input": {
        "query": "machine learning tutorial",
        "type": "",
        "safesearch": false
    }
}

Tool Result:
{
    "results": [
        {
            "title": "Machine Learning Tutorial for Beginners",
            "url": "https://youtube.com/watch?v=abc123",
            "channel": {"name": "ML Course", "id": "UC123"},
            "views": 1500000,
            "duration_formatted": "45:30"
        },
        {
            "title": "Complete ML Course - Full Tutorial",
            "url": "https://youtube.com/watch?v=def456",
            "channel": {"name": "AI Academy", "id": "UC456"},
            "views": 800000,
            "duration_formatted": "12:30:00"
        }
    ]
}
        \end{verbatim}
        
        \item \textbf{Step 1: Postcondition Verification}
        \begin{verbatim}
Postcondition Q_t:
  structural: has_field("results") && is_list(r_t.results)
  semantic: forall video in r_t.results: 
            has_field(video, "title") && has_field(video, "url")

Verification Result: success 
  - Structural check: "results" field exists and is a list 
  - Semantic check: All videos have "title" and "url" fields 
        \end{verbatim}
        
        \item \textbf{Step 1: State Update}
        \begin{verbatim}
S_1 = Update_t(S_0, r_t) = {
    "query": "Find me a tutorial video about machine learning on YouTube",
    "topic": "machine learning",
    "platform": "YouTube",
    "content_type": "tutorial video",
    "youtube_results": [
        {"title": "Machine Learning Tutorial for Beginners", ...},
        {"title": "Complete ML Course - Full Tutorial", ...}
    ],
    "result_count": 19
}
        \end{verbatim}
        
        \item \textbf{Step 2: LLM Reasoning (with tool result)}
        \begin{verbatim}
Conversation History: [
    {"role": "user", "content": "Find me a tutorial video..."},
    {"role": "assistant", "content": "I need to search...\n<start_call_tool>...
    <end_call_tool>"},
    {"role": "user", "content": "<start_tool_result>\n{\"results\": [...]}\n
    <end_tool_result>"}
]

LLM Response:
"Great! I found several machine learning tutorial videos on YouTube.

Based on the YouTube search results, I found several machine learning 
tutorial videos for you:

1. \"Machine Learning Tutorial for Beginners\" - 1.5M views, 45:30
   Channel: ML Course
   URL: https://youtube.com/watch?v=abc123

2. \"Complete ML Course - Full Tutorial\" - 800K views, 12:30:00
   Channel: AI Academy
   URL: https://youtube.com/watch?v=def456

These tutorials cover fundamental concepts and practical applications 
of machine learning."

Action: answer
        \end{verbatim}
        
        \item \textbf{Final State}
        \begin{verbatim}
S_final = {
    "query": "Find me a tutorial video about machine learning on YouTube",
    "topic": "machine learning",
    "platform": "YouTube",
    "content_type": "tutorial video",
    "youtube_results": [...],
    "result_count": 19
}
        \end{verbatim}
    \end{itemize}
\end{mybox}
}


\onecolumn{
\begin{mybox}{Tool Calling Procedure with Postcondition Failure}
    \begin{itemize}
        \item \textbf{Query:} ``What's the weather like in New York today?''
        
        \item \textbf{Step 0: Initial State}
        \begin{verbatim}
S_0 = {
    "query": "What's the weather like in New York today?",
    "location": "New York",
    "date": "today"
}
        \end{verbatim}
        
        \item \textbf{Step 1: LLM Reasoning}
        \begin{verbatim}
Conversation History: []
LLM Response:
"I need to get the current weather information for New York. 
Let me use a weather tool to fetch this data.

<start_call_tool>
Get weather information for New York
<end_call_tool>"

Action: start_call_tool
        \end{verbatim}
        
        \item \textbf{Step 1: Tool Retrieval}
        \begin{verbatim}
Retrieved candidates (top-5 after reranking):
1. Weather_API_GetCurrentWeather (score: 0.91)
2. OpenWeatherMap_CurrentWeather (score: 0.88)
3. WeatherService_GetForecast (score: 0.82)
4. ClimateData_GetWeather (score: 0.75)
5. WeatherInfo_Query (score: 0.71)
        \end{verbatim}
        
        \item \textbf{Step 1: Precondition Checking (Tool 1)}
        \begin{verbatim}
Checking Weather_API_GetCurrentWeather:
  Precondition: exists(location)
  State S_0: {"location": "New York", ...}
  Check: exists(location) = true 
        \end{verbatim}
        
        \item \textbf{Step 1: Tool Execution (Tool 1)}
        \begin{verbatim}
Tool: Weather_API_GetCurrentWeather
Params: {"location": "New York"}
Tool Config: {
    "category": "Weather",
    "api_name": "GetCurrentWeather",
    "tool_input": {"location": "New York"}
}

Tool Result:
{
    "status": "success",
    "data": {
        "city": "New York",
        "temperature": 72,
        "condition": "sunny"
    }
}
        \end{verbatim}
        
        \item \textbf{Step 1: Postcondition Verification (Tool 1)}
        \begin{verbatim}
Postcondition Q_t:
  structural: has_field("temperature") && 
              has_field("condition") &&
              has_field("humidity")
  semantic: temperature is numeric && 
            condition is string

Verification Result: failure 
  - Structural check: Missing required field "humidity" 
  - Semantic check: temperature and condition are valid 
  
Error: Tool result does not contain required field "humidity"
        \end{verbatim}
        
        \item \textbf{Step 1: Tool Marked as Failed}
        \begin{verbatim}
Failed tools: {Weather_API_GetCurrentWeather}
State remains: S_0 (no update)
        \end{verbatim}
        
        \item \textbf{Step 1: Try Next Tool - Precondition Checking (Tool 2)}
        \begin{verbatim}
Checking OpenWeatherMap_CurrentWeather:
  Precondition: exists(location)
  State S_0: {"location": "New York", ...}
  Check: exists(location) = true 
        \end{verbatim}
        
        \item \textbf{Step 1: Tool Execution (Tool 2)}
        \begin{verbatim}
Tool: OpenWeatherMap_CurrentWeather
Params: {"location": "New York"}
Tool Config: {
    "category": "Weather",
    "api_name": "CurrentWeather",
    "tool_input": {"location": "New York"}
}

Tool Result:
{
    "temperature": 72,
    "condition": "sunny",
    "humidity": 65,
    "wind_speed": 10,
    "pressure": 1013,
    "location": "New York, NY"
}
        \end{verbatim}
        
        \item \textbf{Step 1: Postcondition Verification (Tool 2)}
        \begin{verbatim}
Postcondition Q_t:
  structural: has_field("temperature") && 
              has_field("condition") &&
              has_field("humidity")
  semantic: temperature is numeric && 
            condition is string &&
            humidity is numeric

Verification Result: success 
  - Structural check: All required fields present 
  - Semantic check: All fields have correct types 
        \end{verbatim}
        
        \item \textbf{Step 1: State Update}
        \begin{verbatim}
S_1 = Update_t(S_0, r_t) = {
    "query": "What's the weather like in New York today?",
    "location": "New York",
    "date": "today",
    "temperature": 72,
    "condition": "sunny",
    "humidity": 65,
    "wind_speed": 10,
    "pressure": 1013
}
        \end{verbatim}
        
        \item \textbf{Step 2: LLM Reasoning (with tool result)}
        \begin{verbatim}
Conversation History: [
    {"role": "user", "content": "What's the weather like in New York today?"},
    {"role": "assistant", "content": "I need to get the current weather...\n
    <start_call_tool>...<end_call_tool>"},
    {"role": "user", "content": "<start_tool_result>\n{\"temperature\": 72, 
    \"condition\": \"sunny\", ...}\n<end_tool_result>"}
]

LLM Response:
"The weather in New York today is sunny with a temperature of 72°F. 
The humidity is 65%, wind speed is 10 mph, and atmospheric pressure 
is 1013 hPa. It's a pleasant day for outdoor activities."

Action: answer
        \end{verbatim}
        
        \item \textbf{Final State}
        \begin{verbatim}
S_final = {
    "query": "What's the weather like in New York today?",
    "location": "New York",
    "date": "today",
    "temperature": 72,
    "condition": "sunny",
    "humidity": 65,
    "wind_speed": 10,
    "pressure": 1013
}
        \end{verbatim}
    \end{itemize}
\end{mybox}
}

\onecolumn{
\section{Formal Derivations with Hoare Logic and First-Order Contracts}
\label{app:formal-derivations}

In this appendix, we present several representative derivations that make the
logical foundations of ToolGate explicit. We formalize single-step tool
execution, trajectory-level safety, and invariants as Hoare-style proof
obligations and first-order logic (FOL) formulas over the symbolic state space
$\Sigma$ and execution trajectories.

\subsection{Notation and Basic Setting}

We recall that the trusted symbolic state is a typed key--value mapping
$S \in \Sigma$, where
\[
  \Sigma \;=\; \{(k,v,\sigma)\} .
\]
We write $S \models \varphi$ to denote that a (first-order) state formula
$\varphi$ is true in $S$. A tool $t$ is associated with a Hoare-style contract
\[
  \{P_t\} \; t \; \{Q_t\},
\]
where $P_t(S)$ is a state predicate (precondition) and $Q_t(S,r_t)$ is a
postcondition predicate over the pre-state $S$ and runtime result $r_t$:
\[
  Q_t : \Sigma \times R_t \rightarrow \{\text{true}, \text{false}\}.
\]
We write $(S,r_t) \models Q_t$ as shorthand for $Q_t(S,r_t)=1$.

To decouple logical validation from state construction, we introduce a
deterministic state update operator
\[
  \mathsf{Update}_t : \Sigma \times R_t \rightarrow \Sigma,
\]
which specifies the new trusted symbolic state produced when a valid result
$r_t$ is integrated into $S$.

We say that the (ideal) runtime executor of tool $t$ is a (possibly partial)
function
\[
  \mathsf{Exec}(t,S) = r_t
\]
that returns a runtime result $r_t$ when $t$ is invoked under state $S$.

\subsection{Single-Step Hoare-Style Derivation}

We first spell out the standard Hoare-style proof obligation for a single tool
invocation in our setting.

\paragraph{Single-step soundness obligation.}
A contract $\{P_t\}t\{Q_t\}$ is sound w.r.t.\ $\mathsf{Exec}$ and
$\mathsf{Update}_t$ if the following FOL formula holds:
\begin{equation}
\label{eq:single-step-soundness}
  \forall S,r_t.\;
  \Big(
     S \models P_t
     \;\wedge\;
     r_t = \mathsf{Exec}(t,S)
     \;\wedge\;
     Q_t(S,r_t)
  \Big)
  \;\Rightarrow\;
  \mathsf{GoodState}\big(\mathsf{Update}_t(S,r_t)\big),
\end{equation}
where $\mathsf{GoodState}$ expresses that the updated state is well-typed and
consistent (e.g., satisfies global invariants such as key uniqueness and type
soundness).

\paragraph{Inference rule for a single ToolGate step.}
We can capture the operational step of ToolGate for a single tool call as the
following Hoare-style derivation rule:
\[
\infer[\textsc{Tool-Step}]{
  \{\,P_t(S) \wedge \mathsf{Inv}(S)\,\}
  \; t \;
  \{\,\mathsf{Inv}(S') \wedge Q_t(S,r_t) \wedge S' = \mathsf{Update}_t(S,r_t)\,\}
}{
  S \models P_t
  \quad
  r_t = \mathsf{Exec}(t,S)
  \quad
  Q_t(S,r_t)
  \quad
  \mathsf{Inv}(S) \Rightarrow \mathsf{Inv}\big(\mathsf{Update}_t(S,r_t)\big)
}
\]
where $\mathsf{Inv}$ is any chosen state invariant (e.g., that $S$ only contains
verified tool results).

In small-step transition form, a single ToolGate step can be written as
\[
  \langle S_k, R_k \rangle
  \;\xrightarrow{t,\,r_t}\;
  \langle S_{k+1}, R_{k+1} \rangle
\]
with the following proof tree:
\[
\infer[\textsc{Tool-Exec}]{
  \langle S_k, R_k \rangle \xrightarrow{t,r_t} \langle S_{k+1}, R_{k+1} \rangle
}{
  S_k \models P_t
  \quad
  r_t = \mathsf{Exec}(t,S_k)
  \quad
  Q_t(S_k,r_t)
  \quad
  S_{k+1} = \mathsf{Update}_t(S_k,r_t)
  \quad
  R_{k+1} = R_k \cdot \langle t,r_t \rangle
}
\]

\subsection{Precondition Filtering as Weakest Precondition}

Tool selection in ToolGate is constrained by the precondition $P_t$.
We can express this in terms of weakest preconditions. Let
$\mathsf{wp}(t,\Phi)$ be the weakest precondition of tool $t$ w.r.t.\ a desired
post-state formula $\Phi(S')$. Then:
\[
  \mathsf{wp}(t,\Phi)(S)
  \;\; \triangleq \;\;
  \exists r_t.\;
    \Big(
      S \models P_t
      \wedge r_t = \mathsf{Exec}(t,S)
      \wedge Q_t(S,r_t)
      \wedge \Phi\big(\mathsf{Update}_t(S,r_t)\big)
    \Big).
\]

In particular, requiring that $t$ is \emph{executable} in $S$ corresponds to
\[
  S \models \mathsf{wp}(t,\top)
  \;\; \Longleftrightarrow \;\;
  \exists r_t.\;
    S \models P_t \wedge r_t = \mathsf{Exec}(t,S) \wedge Q_t(S,r_t).
\]

The ToolGate precondition filter can then be expressed as:
\begin{equation}
\label{eq:pre-gate}
  \forall t \in C_k.\;
  \mathsf{Admissible}(t,S_k)
  \;\triangleq\;
  S_k \models \mathsf{wp}(t,\top).
\end{equation}

\subsection{Postcondition as Acceptance Event}

The runtime acceptance predicate $A_t$ in ToolGate is defined by:
\[
  A_t(S_k,r_t) \;=\;
  \big(Q_t(S_k,r_t) \wedge \mathsf{wf}(r_t)\big),
\]
where $\mathsf{wf}$ encodes structural and formatting well-formedness for
$r_t$.

We define the state update rule as
\[
  S_{k+1} =
  \begin{cases}
    \mathsf{Update}_t(S_k,r_t) & \text{if } A_t(S_k,r_t) = 1, \\
    S_k                         & \text{otherwise}.
  \end{cases}
\]

This rule can be captured by the following Hoare triple:
\begin{equation}
\label{eq:post-gate}
  \{\,S_k \models P_t\,\}
  \; t \;
  \{\,A_t(S_k,r_t) = 1 \Rightarrow
      \big(S_{k+1} = \mathsf{Update}_t(S_k,r_t) \wedge Q_t(S_k,r_t)\big)\,\}.
\end{equation}

Equivalently, in FOL:
\begin{align}
  \forall S_k,r_t,S_{k+1}.\;
  & S_k \models P_t
    \;\wedge\;
    r_t = \mathsf{Exec}(t,S_k)
    \;\wedge\;
    A_t(S_k,r_t) = 1
  \nonumber\\
  & \Rightarrow\;
    \big(
      S_{k+1} = \mathsf{Update}_t(S_k,r_t)
      \wedge
      Q_t(S_k,r_t)
      \wedge
      \mathsf{GoodState}(S_{k+1})
    \big).
\end{align}

\subsection{Trajectory-Level Safety Derivation}

A full ToolGate execution induces a trajectory
\[
  \tau = \big((S_0,R_0),(t_0,r_0,A_0),\dots,(S_n,R_n)\big).
\]

\paragraph{Per-step safety.}
We say that step $k$ is safe iff:
\begin{align}
\label{eq:step-safe}
\mathsf{SafeStep}_k(\tau)
\triangleq
\Big(
& S_k \models P_{t_k}
\;\wedge\;
  r_k = \mathsf{Exec}(t_k,S_k)
\;\wedge\;
  A_{t_k}(S_k,r_k)=1
\nonumber\\
&\Rightarrow
\big(
  Q_{t_k}(S_k,r_k)
  \wedge
  S_{k+1}
  = \mathsf{Update}_{t_k}(S_k,r_k)
\big)
\Big).
\end{align}

\paragraph{Global safety.}
Trajectory-level safety is then:
\begin{equation}
\label{eq:traj-safe-def}
  \mathsf{Safe}(\tau) \;\triangleq\;
  \bigwedge_{k=0}^{n-1} \mathsf{SafeStep}_k(\tau).
\end{equation}

\paragraph{Soundness theorem (sketch).}
If all tool contracts are sound (Eq.~\ref{eq:single-step-soundness}) and the
initial state $S_0$ satisfies the global invariant $\mathsf{Inv}$, then every
reachable ToolGate trajectory is safe:
\begin{equation}
\label{eq:traj-soundness}
  \forall \tau.\;
  \mathsf{Reach}(q,H,\tau) \wedge S_0 \models \mathsf{Inv}
  \;\Rightarrow\;
  \mathsf{Safe}(\tau) \wedge \bigwedge_{k=0}^{n} \mathsf{Inv}(S_k).
\end{equation}

This can be proved by induction on $k$ using the \textsc{Tool-Step} rule:
\[
\infer[\textsc{Induction}]{
  \forall k.\; \mathsf{Reach}_k(q,H,\tau) \Rightarrow
  \mathsf{SafeStep}_k(\tau) \wedge \mathsf{Inv}(S_k)
}{
  \mathsf{Inv}(S_0)
  \quad
  \forall k.\; \mathsf{SafeStep}_k(\tau) \wedge \mathsf{Inv}(S_k)
  \Rightarrow \mathsf{SafeStep}_{k+1}(\tau) \wedge \mathsf{Inv}(S_{k+1})
}
\]

\subsection{Contract Instantiation for a Concrete Tool}

To illustrate, consider a (simplified) repository management tool
$\mathsf{ListFiles}$ with contract:
\[
  \{P_{\mathsf{list}}\}\;\mathsf{ListFiles}\;\{Q_{\mathsf{list}}\}.
\]

Let the symbolic state contain a key ``\texttt{cwd}'' for the current working
directory and a key ``\texttt{fs}'' for a symbolic file-system abstraction.
We instantiate:
\begin{align}
  P_{\mathsf{list}}(S) &\triangleq
  \exists d.\;
    \big(d = S[\texttt{cwd}] \wedge d \in \mathsf{Dom}(S[\texttt{fs}])\big) ,
  \\
  Q_{\mathsf{list}}(S,r) &\triangleq
  \exists d,L.\;
    d = S[\texttt{cwd}]
    \wedge L = \mathsf{LookupDir}(S[\texttt{fs}],d)
    \wedge r = L,
\end{align}
and define the corresponding state update operator as
\[
  \mathsf{Update}_{\mathsf{list}}(S,r)
  \;=\;
  S \cup \{(\texttt{last\_ls},r,\mathsf{ListType})\}.
\]

The corresponding Hoare triple for this tool is:
\[
  \{\,P_{\mathsf{list}}(S)\,\}\;
  \mathsf{ListFiles}\;
  \left\{\,
    Q_{\mathsf{list}}(S,r)
    \wedge
    S' = \mathsf{Update}_{\mathsf{list}}(S,r)
    \wedge
    \mathsf{Inv}(S')
  \,\right\}.
\]

\paragraph{FOL derivation of a safe call.}
Assume we are at step $k$ with state $S_k$ such that
\[
  S_k \models P_{\mathsf{list}}.
\]
The concrete call is:
\[
  r_k = \mathsf{Exec}(\mathsf{ListFiles},S_k).
\]
Postcondition checking and acceptance give:
\begin{align}
(S_k,r_k) \models Q_{\mathsf{list}}
\;\Rightarrow\;
\exists d,L.\;\nonumber
&\big(d = S_k[\texttt{cwd}]\big)
\\ \nonumber
&\wedge \big(L =
\mathsf{LookupDir}(S_k[\texttt{fs}],d)\big)
\\ \nonumber
&\wedge \big(r_k = L\big),
\\
A_{\mathsf{list}}(S_k,r_k)=1
\;\Rightarrow\;
&S_{k+1}
  = \mathsf{Update}_{\mathsf{list}}(S_k,r_k)
  = S_k \cup \{(\texttt{last\_ls},r_k,\mathsf{ListType})\},
\end{align}
which together imply that $S_{k+1}$ is a well-formed extension of $S_k$.

Combining these, the \textsc{Tool-Exec} rule instantiates to:
\[
\infer[\textsc{Tool-Exec-List}]{
  \langle S_k, R_k \rangle
  \xrightarrow{\mathsf{ListFiles},\,r_k}
  \langle S_{k+1}, R_{k+1} \rangle
}{
\begin{aligned}
  &S_k \models P_{\mathsf{list}}
  \quad
  r_k = \mathsf{Exec}(\mathsf{ListFiles},S_k)
  \\
  &(S_k,r_k) \models Q_{\mathsf{list}}
  \quad
  S_{k+1} = \mathsf{Update}_{\mathsf{list}}(S_k,r_k)
  \quad
  R_{k+1} = R_k \cdot \langle \mathsf{ListFiles},r_k \rangle
\end{aligned}
}
\]

\subsection{Contract-Governed Tool Selection Policy}

Finally, we combine the probabilistic ranking distribution with logical
filtering. Let $\mathsf{rank}(t \mid u_k)$ be the (normalized) ranking score
over candidate tools given requirement representation $u_k$.

We define the \emph{contract-governed policy}:
\[
  \pi(t \mid q,H,S_k,R_k)
  \;\triangleq\;
  \frac{
    \mathsf{rank}(t \mid u_k) \cdot \mathbf{1}[S_k \models P_t]
  }{
    \sum_{t' \in C_k} \mathsf{rank}(t' \mid u_k)\cdot \mathbf{1}[S_k \models P_{t'}]
  }.
\]

A trajectory $\tau$ is then sampled according to:
\begin{align}
  p(\tau \mid q,H)
  \;=\;
  \prod_{k=0}^{n-1}
  &\Big(
    p(\langle S_k,R_k\rangle)
    \cdot
    p(\text{\texttt{<start\_call\_tool>}} \mid q,H,S_k,R_k)
    \cdot
    \pi(t_k \mid q,H,S_k,R_k)
  \nonumber\\[-0.3em]
  &\cdot\;
    p(r_k = \mathsf{Exec}(t_k,S_k))
    \cdot
    p(A_{t_k}(S_k,r_k)=1 \mid S_k,r_k)
  \Big),
\end{align}
subject to the global constraint that any violation of $P_t$ or $Q_t$ yields
zero probability:
\[
  \exists k.\;
  \neg \mathsf{SafeStep}_k(\tau)
  \;\Rightarrow\;
  p(\tau \mid q,H) = 0.
\]

This explicit factorization makes the interaction between probabilistic
reasoning and logical contracts formally visible and verifiable.

}

\end{document}